\documentclass[journal]{IEEEtran}
\usepackage{amsmath,amsfonts}
\usepackage{algorithmic}
\usepackage{algorithm}
\usepackage{array}
\usepackage{caption}
\DeclareCaptionLabelSeparator{newline}{\par}
\usepackage{textcomp}
\usepackage{stfloats}
\usepackage{url}
\usepackage{verbatim}
\usepackage{graphicx}
\usepackage{subcaption}
\usepackage{cite}
\usepackage{booktabs}

\begin{document}

\title{SIESEF-FusionNet: Spatial Inter-correlation Enhancement and Spatially-Embedded Feature Fusion Network for LiDAR Point Cloud Semantic Segmentation}

\author{Jiale Chen, Fei Xia,~\IEEEmembership{Member,~IEEE,} Jianliang Mao,~\IEEEmembership{Member,~IEEE,} Haoping Wang, \\and Chuanlin Zhang,~\IEEEmembership{Senior Member,~IEEE}
\thanks{Jiale Chen, Fei Xia, Jianliang Mao, and Chuanlin Zhang are with the Intelligent Autonomous Systems Laboratory, Shanghai University of Electric Power, Shanghai 200090, China (e-mail: cjlchen@mail.shiep.edu.cn; xiafei@shiep.edu.cn; jl\_mao@shiep.edu.cn; clzhang@shiep.edu.cn). \textit{(Corresponding author: Fei Xia.)}{\tt\small }}
\thanks{Haoping Wang is with the College of Literature, Science and the Arts, University of Michigan-Ann Arbor, Ann Arbor, MI 48109, USA (e-mail: haopingw@umich.edu).{\tt\small }}}

\markboth{Journal of \LaTeX\ Class Files,~Vol.~14, No.~8, August~2021}%
{Shell \MakeLowercase{\textit{et al.}}: A Sample Article Using IEEEtran.cls for IEEE Journals}

\maketitle

\begin{abstract}

The ambiguity at the boundaries of different semantic classes in point cloud semantic segmentation often leads to incorrect decisions in intelligent perception systems, such as autonomous driving. Hence, accurate delineation of the boundaries is crucial for improving safety in autonomous driving. A novel spatial inter-correlation enhancement and spatially-embedded feature fusion network (SIESEF-FusionNet) is proposed in this paper, enhancing spatial inter-correlation by combining inverse distance weighting and angular compensation to extract more beneficial spatial information without causing redundancy. Meanwhile, a new spatial adaptive pooling module is also designed, embedding enhanced spatial information into semantic features for strengthening the context-awareness of semantic features. Experimental results demonstrate that 83.7\% $mIoU$ and 97.8\% $OA$ are achieved by SIESEF-FusionNet on the Toronto3D dataset, with performance superior to other baseline methods. A value of 61.1\% $mIoU$ is reached on the semanticKITTI dataset, where a marked improvement in segmentation performance is observed. In addition, the effectiveness and plug-and-play capability of the proposed modules are further verified through ablation studies.

\end{abstract}


\section{Introduction}
\IEEEPARstart{L}{iDAR} point cloud semantic segmentation is considered crucial for scene understanding. It is regarded as the foundation for safe navigation in autonomous driving \cite{sharma2023real} and intelligent robotics \cite{li2023autonomous}. However, due to the unstructured, unordered nature and sparsity of LiDAR point cloud data, traditional 2D deep learning methods \cite{Peng2024} are difficult to directly apply to 3D point cloud processing \cite{ibrahim2023sat3d}.
In recent years, research in point cloud semantic segmentation has primarily focused on transforming point clouds into more regular structures, with projection-based \cite{su2015multi,tatarchenko2018tangent,wen2022hybrid} and voxel-based \cite{maturana2015voxnet,tchapmi2017segcloud,zhu2021cylindrical} methods included. However, the loss of critical local spatial features and the increase in computational overhead are accompanied by such methods during the transformation process \cite{li2024tpv}.
Recently, researchers have proposed methods for the direct processing of raw point cloud data \cite{qi2017pointnet,qi2017pointnet++}, which reduce information loss, minimize dependence on complex processing and improve computational efficiency. And to overcome the difficulties associated with processing large-scale outdoor LiDAR point cloud data, hierarchical sampling methods and various local feature encoding modules have been proposed \cite{qi2017pointnet++,hu2020randla,qiu2021semantic}. The enhancement of local contextual features in point clouds is improved by these approaches.

Despite improvements in segmentation performance, several challenges are still encountered by point-based methods: 1) Boundary overlaps between different semantic classes are often unavoidable during neighborhood construction. Spatial features in existing methods are either inadequately encoded due to improper selection of spatial information or rigidly integrated in a mechanical manner. The inter-correlation of spatial information is inadequately considered in such integration, leading to information redundancy. 2) The connectivity of point cloud boundary points is weakened, and the accuracy of boundary segmentation is reduced due to insufficient fusion of spatial and semantic information. In applications of autonomous driving and robot navigation, potential risks are met by this boundary ambiguity, directly impacting navigation decisions and the safety of perception systems. 

To tackle the issue of boundary ambiguity between different semantic classes, a LiDAR point cloud semantic segmentation network SIESEF-FusionNet is proposed, based on enhanced spatial interconnectivity and spatially-embedded feature fusion. Specifically, an enhanced local spatial encoding (ELSE) module is designed to strengthen spatial inter-correlation by incorporating inverse distance weighting and angular compensation information. Additionally, spatially-embedded adaptive pooling (SEAP) module is utilized to embed the enhanced local spatial encoding into local semantic features for improving the contextual representation capability of local features. The proposed method offers the following advantages over existing point-based approaches: 1) Effectiveness: the impact of point cloud boundary ambiguity is mitigated and enhancing semantic segmentation performance; 2) Plug-and-play capability: integration into various model architectures can be readily achieved, effectively enhancing the feature learning capability of the model. The competitive segmentation performance of SIESEF-FusionNet is evaluated through extensive experiments on two outdoor LiDAR datasets, demonstrating the potential of the proposed method for intelligent transportation scenarios. Furthermore, the effectiveness and plug-and-play capability of the proposed modules are further validated through ablation studies. In summary, the primary contributions of this work are as follows:
\begin{itemize}
\item A novel ELSE module is proposed to enhance the inter-correlation of local spatial information, thereby mitigating boundary ambiguity in the semantic segmentation of point clouds. To the best of our knowledge, this work is the first to consider spatial inter-correlation by utilizing inverse distance weighting and angular compensation in the application of LiDAR semantic segmentation.
\item A SEAP module is devised to embed enhanced local spatial encoding within local semantic features, preserving more features with sufficiently combined spatial information and semantic information than other pooling methods, thereby augmenting contextual perception capabilities.
\item Compared to the existing methods, the proposed ELSE and SEAP modules are plug-and-play compatible, allowing for effortless integration into point cloud segmentation networks and enhancing deep feature extraction.
\end{itemize}

The remainder of this paper is organized as follows. Section II reviews recent advancements in 3D semantic segmentation. Section III provides a comprehensive description of SIESEF-FusionNet. Section IV provides a detailed account of extensive experiments and analyzes the results. Finally, Section V summarizes the main conclusions.

\section{Related Work}
\subsection{Deep Learning-based Methods}
Recent deep learning methods for point cloud semantic segmentation are divided into projection-based, voxel-based, and point-based approaches. Projection-based semantic segmentation methods \cite{su2015multi,tatarchenko2018tangent,wen2022hybrid,zhang2020polarnet,xu2020squeezesegv3,milioto2019rangenet++} are applied by projecting the 3D point clouds onto 2D images for subsequent processing. Conversely, voxel-based methods \cite{maturana2015voxnet,tchapmi2017segcloud,zhu2021cylindrical,hou2022point,mi2021automated} are utilized to transform 3D point clouds into regular 3D voxels. Both approaches are designed to convert unstructured point cloud data into either structured 2D images or 3D voxels, which facilitates the application of 2D CNNs or 3D CNNs for processing. Nevertheless, both projection and voxelization methods result in the loss of geometric information and incur substantial computational costs.
To alleviate the issues caused by projection and voxelization, point-based methods \cite{qi2017pointnet,qi2017pointnet++,qiu2021geometric,thomas2019kpconv,hu2020randla,qiu2021semantic,xu2023neiea} and \cite{luo2023dense} are utilized for directly learning features from raw point clouds. Consequently, this approach reduces information loss during data structure conversion and enhances computational efficiency. A shared MLP applied independently to each point in the point cloud for learning point-wise features is proposed in \cite{qi2017pointnet}, while a hierarchical learning architecture building on \cite{qi2017pointnet} to capture local features is proposed in \cite{qi2017pointnet++}. The successes of these two methods emphasize that directly extracting features from raw point clouds is superior to traditional projection-based and voxel-based approaches. Consequently, point-based methods have emerged as the predominant approach for 3D point cloud semantic segmentation. Point-based methods have been developed, including frameworks such as PointNet, PointNet++, RandLA-Net, and U-NEXT, and the framework adopted in this paper is U-NEXT \cite{zeng2023small}.
\subsection{Encoding of Local Context Features}
The precise capture and analysis of local context information are essential for improving the quality, accuracy, and reliability of point cloud semantic segmentation. In recent years, numerous methods have been proposed for local feature extraction. A CNN module incorporating an error-correction feedback mechanism is proposed in \cite{qiu2021geometric}, enabling the efficient learning of local features in point clouds. Concurrently, explicitly defined kernel points are utilized in \cite{thomas2019kpconv} for scalable convolution, allowing local features to be captured and improving computational efficiency. Nonetheless, there are some limits in the above methods when applied to outdoor LiDAR scenes, with the reason that considerable variations in point cloud density and pronounced sparsity are observed in outdoor LiDAR scenes \cite{zhang2021surrf}. To tackle this issue, various hierarchical downsampling techniques are utilized, as demonstrated in \cite{hu2020randla} and \cite{qiu2021semantic}, to mitigate time complexity. Nevertheless, multiple downsampling operations result in the loss of local detail, leading to ambiguity at the boundaries between different semantic classes in the point cloud. To refine the local contextual features, various local feature encoding modules have been proposed in works such as \cite{hu2020randla}, \cite{qiu2021semantic}, \cite{xu2023neiea}, and \cite{luo2023dense}. Nevertheless, the overlap of boundaries between different semantic classes is an inherent issue during neighborhood construction. Limits are often encountered when processing large-scale outdoor LiDAR data, including marked by complex scenes, uneven point cloud density, and sparse distributions. Such issues arise due to the insufficient consideration of spatial inter-correlation. Consequently, an ELSE module is proposed to improve the spatial inter-correlation among local positional information, as detailed in Section III-B.
\subsection{Aggregation of Local Context Features}
Integrating local spatial positions with semantic information represents a crucial aspect in enhancing the performance of point cloud semantic segmentation. \cite{qi2017pointnet++} employs max pooling to aggregate local features of point clouds. \cite{hu2020randla} and \cite{fan2021scf} utilize attention-based pooling strategies for the adaptive aggregation of features. Nevertheless, both local salient features and local detailed features are not extracted simultaneously. \cite{qiu2021semantic} and \cite{zeng2022leard} combine max-pooled features with attention-based features using a straightforward concatenation fusion approach. However, important local contextual information may be neglected by this approach. The limits of the aforementioned pooling methods result in an inadequate integration of spatial and semantic information, which fails to preserve the interrelationships at the boundaries, thereby decreasing the accuracy of boundary segmentation. Therefore, a SEAP module is proposed in this paper, which adaptively integrates enhanced local spatial encodings into the semantic features to reinforce their inter-correlation. Further details are provided in Section III-C.

\section{Method}
\subsection{Network Architecture}

\begin{figure*}[!t]
\centering
\includegraphics[scale=0.17]{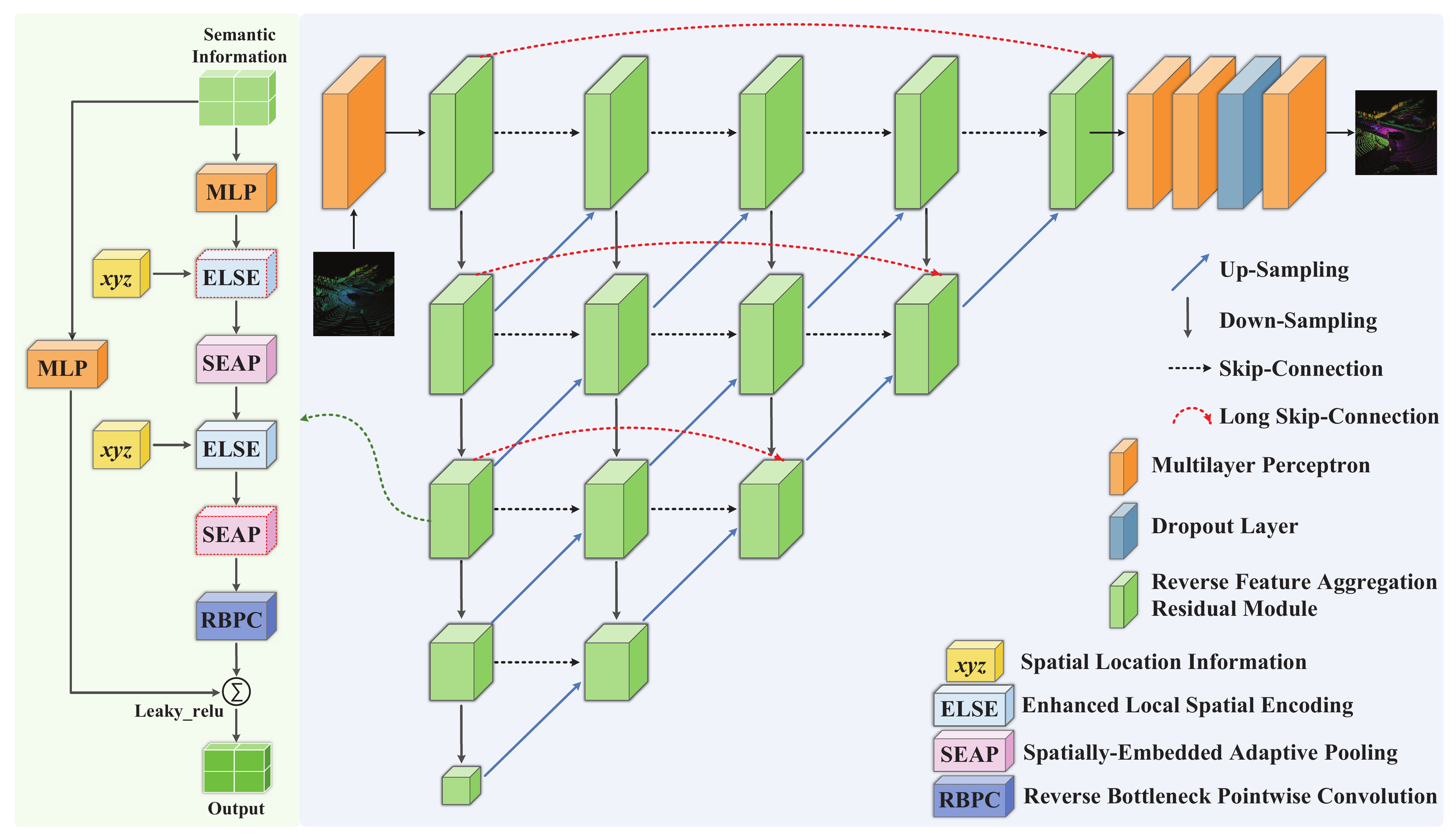}
\caption{The proposed SIESEF-FusionNet architecture.}
\label{fig_1}
\end{figure*}

\begin{figure*}[]
\centering
\includegraphics[scale=0.2]{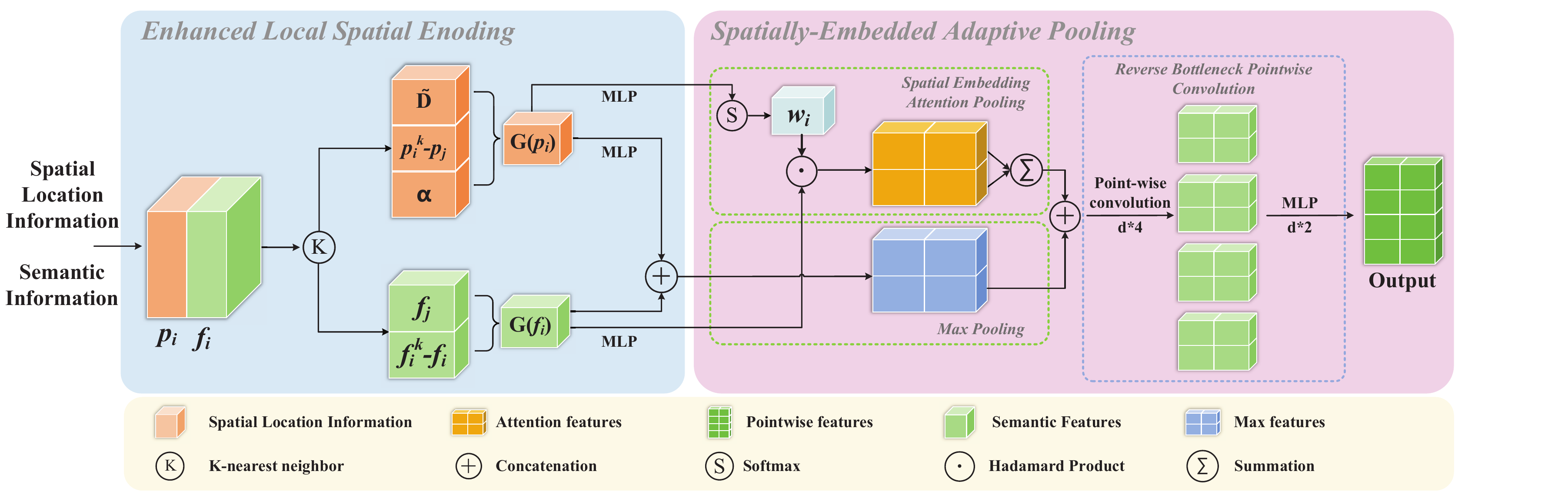}
\caption{The proposed reverse feature aggregation residual module is illustrated with the top part showing connections between these modules and the bottom part detailing the ELSE and SEAP modules.}
\label{fig_2}
\end{figure*}
A novel network SIESEF-FusionNet is proposed to address the problems inherent in semantic segmentation of point clouds within LiDAR environments. Multi-resolution features are extracted by SIESEF-FusionNet using the U-NEXT framework \cite{zeng2023small}, as illustrated in Fig. \ref{fig_1}. The reverse feature aggregation residual module, composed of the ELSE and SEAP modules, is shown on the left side of Fig. \ref{fig_1}. Details of the ELSE and SEAP modules are presented in Fig. \ref{fig_2}.

\subsection{Enhanced Local Spatial Encoding Module}

The inter-correlation of spatial location information is strengthened by the ELSE module through the implementation of inverse distance weighting and angular compensation, both applied to neighboring points with respect to the centroid. A more robust basis for resolving ambiguity at the boundaries between distinct semantic classes in point clouds is established by this improvement.
The formula for enhancing local spatial features $G(p_i)$ is provided as:

\begin{small}\begin{equation}G(p_i)=MLP((p_i^k-p_i)\oplus\widetilde{D}\oplus\alpha(p_i,p_i^k)),\end{equation}\end{small}where $p_i\in\mathbb{R}^3$ is the centroid point, $\{p_i^k:n\in N,1\leq k\leq K\}$ is the $K$ neighboring points of the centroid point, $p_i^k-p_i\in\mathbb{R}^3$ is the relative position of the neighboring points with respect to the centroid point, $\widetilde{D}$ is the inverse distance weight, $\alpha(p_i,p_i^k)$ is the angle compensation information of the neighboring points relative to the centroid point. 

Inverse distance weighting is employed as an adaptive mechanism for weight adjustment. It is based on the principle that the influence of a neighboring point on the centroid decreases as the distance increases. The weight distribution strategy is designed to prioritize spatial positional information by allocating larger weights to points that are closer to the centroid. As a result, the mutual influence between adjacent points is increased, thereby improving the accuracy and reliability of segmentation. The inverse distance weight is defined as follows:

\begin{small}\begin{equation}\widetilde{D}=1-S(D),\end{equation}
\end{small}where $S(\cdot)$ is the softmax function, ${D}$ is the Euclidean distance of the neighboring points relative to the centroid point.

Incorporating angular information optimizes the directionality of spatial positional data, ensuring that boundary points in varying directions receive appropriate attention. The arctangent values of neighboring points relative to the centroid are calculated along the $XYZ$ axes. Subsequently, the corresponding sine values and cosine values are derived. These angular metrics are then integrated into the model as compensation factors to enhance its accuracy, thereby refining the spatial information representation. As a result, the model sensitivity and ability to capture spatial structures are appreciably strengthened. First, the three-dimensional vectors are calculated based on $p_{i}$ and $p_i^k$, as shown below:

\begin{small}\begin{equation}\mathrm{delta}=(\delta_x,\delta_y,\delta_z),\end{equation}\end{small}where $\delta_{x}$,$\delta_{y}$,$\delta_{z}$ are $x_i^k-x_i$,$y_i^k-y_i$,$z_i^k-z_i$, representing the components in the $XYZ$ directions, respectively. Then, the arctangent values of the neighboring points relative to the centroid point in the $XYZ$ directions are calculated, as shown below:

\begin{small}\begin{equation}\begin{cases}\theta(xy)=\arctan(\delta_y,\delta_x)\\\theta(yz)=\arctan(\delta_z,\delta_y)\\\theta(zx)=\arctan(\delta_z,\delta_x).\end{cases}\end{equation}\end{small}

To avoid the discontinuity and numerical stability difficulties associated with the arctangent function, the sine functions and cosine functions are integrated. The inherent ambiguities of the arctangent function, particularly the discontinuous transitions near $\pm\frac{\pi}{2}$, are mitigated by the integration. Sine values and cosine values are utilized to capture more accurate dynamic angle variations, thereby enhancing the robustness of the algorithm. Finally, the normalized sine and cosine values are combined into a vector, and the final angle compensation information is obtained as:

\begin{small}\begin{equation}\alpha(p_i,p_i^k)=[N(\beta(xy),\beta(yz),\beta(zx))],\end{equation}\end{small}where $N(\cdot)$ is the normalization operationa and $\beta(\cdot)=[\sin(\theta(\cdot)),\cos(\theta(\cdot))]$.

\subsection{Spatially-Embedded Adaptive Pooling Module}

The enhanced spatial positional information is integrated as a scaling factor for semantic feature encoding during aggregation by the SEAP module. As a result, the adaptive learning of contextual features within local semantic neighborhoods is enabled.

Firstly, the enhanced spatial adaptive weights are computed using an MLP and the softmax function, illustated as:

\begin{small}\begin{equation}w=\frac{\exp{(MLP(G(p_i)))}}{\sum_{i=1}^K\exp{(MLP(G(p_i)))}}.\end{equation}\end{small}

Subsequently, the enhanced spatial adaptive weights are applied to the local semantic features on a point-by-point basis. The weighted local feature information $\tilde{F}(f_i)$ is yielded through the process, shown as:

\begin{small}\begin{equation}\tilde{F}(f_i)=\sum_{k=1}^K(F(f_i)\odot w),\end{equation}\end{small}where $F(f_i)$ is the local semantic features, shown as:

\begin{small}\begin{equation}F(f_i)=MLP((f_i^k-f_i)\oplus f_i^k),\end{equation}\end{small}where $f_i^k-f_i$ is the relative semantic features, with $f_i^k$ being the neighboring feature.

The enhancement of detail capture and feature expressiveness is achieved through the integration of salient features derived from max pooling. The ability to capture fine-grained details is substantially augmented through the process. The SEAP module is described as:

\begin{small}\begin{equation}F(f_i)=\tilde{F}(f_i)\oplus MAX(F(f_i)\oplus G(p_i)),\end{equation}\end{small}where $MAX(\cdot)$ is max pooling.

To enhance point-wise features, the reverse bottleneck point convolution (RBPC) module is proposed, as shown in Fig. \ref{fig_1}. Initially, the outputs from two SEAP modules are concatenated by the module. Subsequently, feature dimensions are expanded through per-point convolution, and the output feature dimensions are adjusted using an MLP. The receptive field of each point is substantially improved, and the feature extraction hierarchy is deepened through this approach. The process is detailed as:

\begin{small}\begin{equation}F_{Invout}(f_i)=MLP(\phi(F^1(f_i)\oplus F^2(f_i))),\end{equation}\end{small}where $F^1(f_i)$ is the feature output from the first SEAP module, and $F^2(f_i)$ is the feature output from the second SEAP module and $\phi(\cdot)$ is the per-point convolution.

To mitigate the vanishing gradient problem, a residual connection is incorporated between the original input and output features. This connection is activated by the Leaky ReLU function. Additionally, to enhance convergence speed and capture inter-class information, a weighted cross-entropy loss function is employed:

\begin{small}\begin{equation}L=-\frac1N\sum_{i=1}^N\sum_{j=1}^Cw_jy_{ij}log(\hat{y}_{ij}),\end{equation}\end{small}where $N$ is the number of samples, $C$ is the number of classes, $w_{j}$ is is the weight for class $j$, $y_{ij}$ is the true label of the $i$-th sample for class $j$, and $\widehat{y}_{ij}$ is the predicted probability of the  $i$-th sample belonging to class $j$.

\section{Experimental Verification}
Experiments are performed on two large-scale autonomous driving datasets: Toronto3D \cite{tan2020toronto} and SemanticKITTI \cite{behley2019semantickitti}. Initially, both quantitative and visual results from these datasets are presented. Subsequently, these results are compared with existing methods to validate the performance of the proposed network. Finally, a comprehensive ablation study is conducted on the Toronto3D dataset to further analyze the performance and plug-and-play capability of the ELSE and SEAP modules in various network architectures. The variants of the ablation experiments are outlined as follows.
\begin{enumerate}
\item A1 is the baseline model. Basic local spatial information (limited to relative positions) is used to replace ELSE, while max pooling is employed to aggregate semantic and spatial features in this model.
\item In A2, ELSE is utilized to enhance the inter-correlation between local spatial features and to improve encoding conditions.
\item In A3, SEAP is employed to adaptively learn the contextual features of local semantic neighborhoods and their inter-correlations.
\item B1 is the RandLA-Net model.
\item In B2, ELSE and SEAP replace the local spatial encoding and attention pooling modules of RandLA-Net, respectively.
\item B3 is the BAAF-Net model.
\item In B4, ELSE and SEAP replace the bilateral context encoding and hybrid local aggregation modules of BAAF-Net, respectively.
\end{enumerate}

\subsection{Dataset Introduction}

The Toronto3D dataset \cite{tan2020toronto} is collected using a vehicle-mounted Mobile Laser Scanning (MLS) system and covers approximately 1 kilometer of road scenes in Toronto, Canada. Region 2 is designated as the test set, while the other three regions are utilized for training.  The SemanticKITTI dataset \cite{behley2019semantickitti} is featured with complex outdoor traffic scenarios and is composed of 22 densely recorded stereo sequences, with precise annotations provided across 19 semantic classes. The sequences 00 to 07 and 09 to 10 are designated as the training set, with sequence 08 serving as the validation set, while the results from sequences 11 to 21 are utilized for online evaluation.

\begin{table*}[htbp]
\caption{Semantic Segmentation Results on Toronto3D Dataset (\%)}
\label{table_1}
\begin{center}
\resizebox{\textwidth}{!}{
\begin{tabular}{clcccccccccccc}
\toprule
Input  &Method  & OA  & mIoU  & road  & road marking & nature  & building  & utility line  & pole & car  & fence\\
\midrule
& PointNet++\cite{qi2017pointnet++}  & 92.6  & 59.5  & 92.9  & 0.0  & 86.1  & 82.2  & 60.9  & 62.8  & 76.4  & 14.4\\
& KPConv\cite{thomas2019kpconv} & 95.4  & 69.1  & 94.6  & 0.1  & 96.1  & 91.5  & 87.7 
& 81.6  & 85.7  & 15.7\\
& RandLA-Net\cite{hu2020randla}  & 93.0  & 77.7  & 94.6  & 42.6  & 96.9  & 93.0  & 86.5  & 78.1  & 92.9  & 37.1\\
& BAAF-Net\cite{qiu2021semantic}  & 97.1  & \underline{80.9}  & \underline{96.9}  & \underline{68.1}  & 96.1  & 91.2  & 87.1  & 83.0  & 88.9  & 36.0\\
XYZ & LACV-Net\cite{zeng2024large}  & 95.8  & 78.5  & 94.8  & 42.7  & 96.7  & 91.4  & 88.2 
& 79.6  & 93.9  & 40.6\\
& PointNeXt\cite{qian2022pointnext}  & 95.9  & 79.5  & 95.7  & 45.4  & \underline{97.2}  & 93.4  & \underline{88.9}  & 81.5  & 93.7  & 40.4\\
& MVP-Net\cite{li2023mvpnet}  & 96.1  & 75.1  & 95.1  & 22.0  & 96.5  & 92.8  & 88.3  & \underline{85.0}  & 91.8  & 29.5\\
& SFL-Net\cite{li2023sfl}  & 96.0  & 78.1  & 94.2  & 34.0  & 96.9 & \underline{93.8}  & 87.1  & \textbf{85.7}  & 93.5  & 39.7\\
& PointNAT\cite{zeng2024pointnat}  & \underline{96.5}  & 80.6  & 96.2  & 48.5  & \textbf{97.5}  & 93.7  & \textbf{90.1}  & 83.2  & \underline{94.0}  & \underline{41.5}\\
\cmidrule(lr){2-12} 
& Proposed Method  & \textbf{97.8}  & \textbf{83.7}  & \textbf{97.7}  & \textbf{72.3}  & \underline{97.2}  & \textbf{94.3}  & 87.4  & 80.0  & \textbf{95.0}  & \textbf{45.6}\\
\bottomrule 
\end{tabular}}
\end{center}
\end{table*}

\subsection{Implementation Details}
To evaluate the segmentation results by our proposed method, overall accuracy ($OA$), per-class intersection over union ($IoU$), and mean IoU ($mIoU$) are used as evaluation metrics, which are shown below:

\begin{small}\begin{equation}IoU=\frac{TP}{FN+FP+TP},\end{equation}\end{small}
\begin{small}\begin{equation}mIoU=\frac1k\sum_1^k\frac{TP}{FN+FP+TP},\end{equation}\end{small}
\begin{small}\begin{equation}OA=\frac{TP+TN}{TP+TN+FP+FN},\end{equation}\end{small}where $TP$,$TN$,$FP$,$FN$ are true positive, true negative, false positive, and false negative samples, respectively, and $k$ represents the number of classes.

All experiments are conducted using the TensorFlow framework on an NVIDIA RTX 4090 24GB GPU. The Adam optimizer is used for training. The training runs for 100 epochs with the number of neighbor points set to $K=16$. The initial learning rate is set to 0.01 and is reduced by 5\% after each epoch.

\subsection{Evaluation on the Toronto3D Dataset}
A detailed quantitative comparison between the proposed method and various baseline methods on the Toronto3D dataset is presented in Table \ref{table_1}. The best results are indicated in bold, while the second-best results are underlined. The inter-correlation of spatial positions is emphasized by the proposed method, thereby necessitating the specific selection of baseline methods that utilize only $XYZ$ coordinates for input features to ensure a fair comparison. Exceptional performance is exhibited by the proposed algorithm in both $OA$ $\left(97.8\%\right)$ and $mIoU$ $\left(83.7\%\right)$, with $OA$ exceeding RandLA-Net and BAAF-Net by 0.7\% and 4.8\%, respectively, and $mIoU$ surpassing them by 2.8\% and 6\%, respectively. In terms of $IoU$ scores for individual classes, the proposed method achieves superior performance in 5 out of 8 classes, such as road, road markings, buildings, cars, and fences, which are indicated in bold in Table \ref{table_1}. Distinct advantages over other baseline methods are demonstrated by the proposed method in the semantic segmentation of road markings, with an $IoU$ of 72.3\% being achieved in this class. An $IoU$ of 97.2\% is achieved by the proposed method in the segmentation of nature class, which ranks second among the $IoU$ of all baseline methods, with only a 0.3\% difference from the best-performing method. It is noted that better performance is demonstrated by the proposed method in large-scale scene classes, such as nature and buildings, as well as in smaller classes characterized by limited samples and diverse types, such as fences and cars. It is indicated by the results that the proposed method can be effectively adapted to classes of varying scales. The adaptability to classification ability at different scales is attributed to the increased focus of the network on the interrelations of local spatial features, where not only the semantic boundaries of point clouds are refined, but also the ability to capture global contextual features is enhanced, resulting in more precise segmentation inference.

\begin{figure*}[htbp]
  \centering
  \includegraphics[scale=0.32]{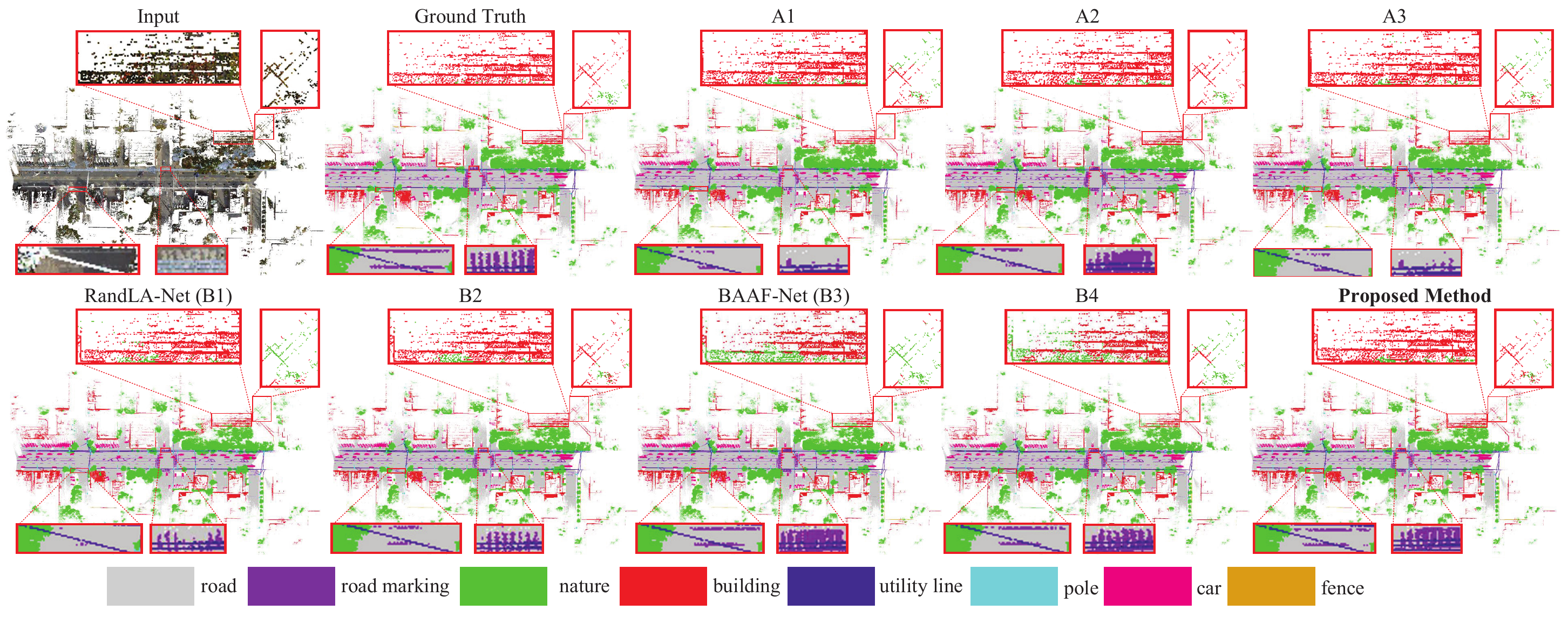}
  \caption{Comparison of semantic segmentation results on the Toronto3D dataset.}
  \label{fig_3}
  \end{figure*}

As shown in Fig. \ref{fig_3}, a visual comparison of results is conducted on the Toronto3D dataset. In addition to the comparison between the visual results of the proposed algorithm and two baseline models, the visual results of the variants from ablation studies are also compared. 
Marked superiority over RandLA-Net and BAAF-Net is demonstrated by the proposed method in the segmentation of natural features, buildings, and road markings. It is noteworthy that in the areas emphasized by red boxes, the boundary segmentation of road markings, particularly for sidewalks, is greatly enhanced by the proposed method. In the segmentation results of RandLA-Net, substantial portions of the road markings are missing, while ambiguity at the boundary between road markings and the road is indicated in the results of BAAF-Net, leading to mis-segmentation of parts of the road as road markings. The reason can be attributed to the fact that both of these methods, along with other baseline approaches, directly integrate spatial information encoding without fully considering the inter-correlation among local spatial positions. As a result, boundary ambiguity and mis-segmentation of the point cloud are induced. In contrast, the inter-correlation among local spatial information are enhanced in the proposed method through the use of inverse distance weighting and angular compensation. The adaptive embedding of this information into the semantic features is facilitated by the SEAP module, which results in clearer and more accurate boundaries of the point cloud and effectively mitigates segmentation errors resulting from incomplete segmentation and those caused by boundary ambiguity. In the visual results of the ablation study, greater accuracy in segmenting road markings and natural features is clearly observed in the segmentation results of the proposed method and its three variant models (A1, A2, A3). The effectiveness of each module in the proposed method is illustrated through the visual comparison of the results in the Group A ablation experiments. In the Group B ablation experiments, the proposed module combinations are applied within two different algorithm frameworks, which are represented by variant models B2 and B4. It is clearly observed that more complete and accurate segmentation of road markings is achieved by the B2 model, compared to the baseline model B1. In the Group B ablation experiments, the proposed module combinations are utilized within two different algorithm frameworks, represented by variant models B2 and B4. It is clearly observed that, despite the relatively limited accuracy improvements achieved by models B2 and B4 in segmenting the boundaries of nature and building, a more complete and accurate segmentation of road markings is accomplished by the B2 model when compared to the baseline model B1. Additionally, enhanced accuracy and smoothness in segmenting both road markings and road boundaries are demonstrated by the B4 model when compared to the baseline model B3. The plug-and-play capability and flexibility of the proposed modules are demonstrated through the comparison of the visualization results from the ablation experiments conducted with models B1 to B4.

\begin{figure*}[htbp]
\centering
\begin{subfigure}[b]{.43\linewidth}
    \centering
    \includegraphics[scale=0.70]{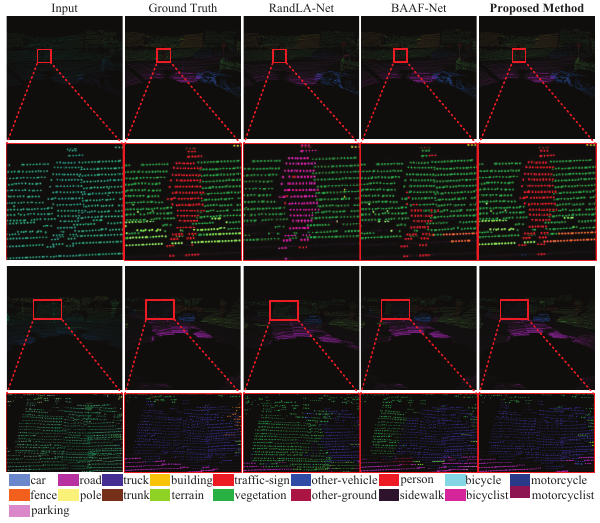}
    \caption{}
\end{subfigure}
\hspace{1cm} 
\begin{subfigure}[b]{.43\linewidth}
    \centering
    \includegraphics[scale=0.70]{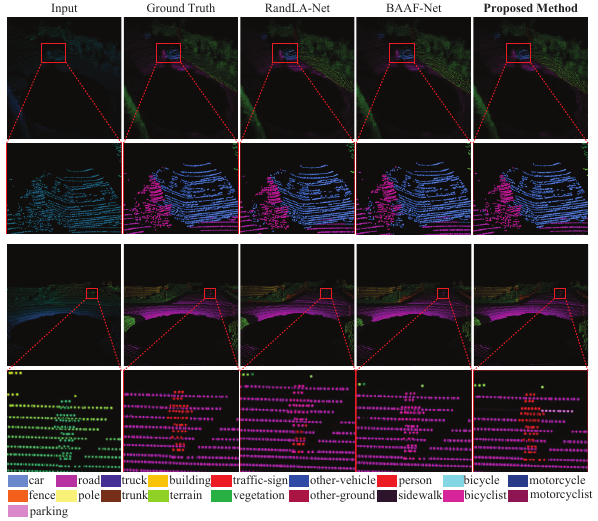}
    \caption{}
\end{subfigure}
\caption{Comparison of semantic segmentation results on the SemanticKITTI dataset. (a) Visualization of segmentation results for persons, trucks, and vegetation. (b) Visualization of segmentation results for bicyclists and cars, persons and roads.}
\label{fig_4}
\end{figure*}
        
  \begin{table*}[htbp]
  \caption{Semantic Segmentation (Single-Scan) Results on the SemanticKITTI Dataset (\%)}
  \label{table_2}
  \begin{center}
  \resizebox{\textwidth}{!}{
  \begin{tabular}{lcccccccccccccccccccccc}
  \toprule
  Method   & \rotatebox{90}{mIoU}  & \rotatebox{90}{road}  & \rotatebox{90}{sidewalk} & \rotatebox{90}{parking}  & \rotatebox{90}{other-ground}  & \rotatebox{90}{building}  & \rotatebox{90}{Car} 
      & \rotatebox{90}{truck}  & \rotatebox{90}{bicycle}  & \rotatebox{90}{motorcycle}  & \rotatebox{90}{other-vehicle}  & \rotatebox{90}{vegetation}  & \rotatebox{90}{trunk}  & \rotatebox{90}{terrain}  & \rotatebox{90}{person}  & \rotatebox{90}{bicyclist}  & \rotatebox{90}{motorcyclist}  & \rotatebox{90}{fence}
      & \rotatebox{90}{pole}  & \rotatebox{90}{traffic-sign} & \rotatebox{90}{score}\\
  \midrule
      PointNet\cite{qi2017pointnet}   & 14.6  & 61.6 & 35.7 & 15.8  & 1.4  & 41.4  & 46.3  & 0.1  & 1.3  & 0.3  & 0.8  & 31.0  & 4.6  & 17.6  & 0.2  & 0.2  & 0.0  & 12.9  & 2.4  & 3.7 & 0\\
      PointNet++\cite{qi2017pointnet++}   & 20.1  & 72.0  & 41.8  & 18.7  & 5.6  & 62.3  & 53.7  & 0.9  & 1.9  & 0.2  & 0.2  & 46.5  & 13.8  & 30.0  & 0.9  & 1.0  & 0.0  & 16.9  & 6.0  & 8.9 & 0\\
      TangentConv\cite{tatarchenko2018tangent} & 40.9 & 83.9  & 63.9  & 33.4  & 15.4  & 83.4  & 90.8  & 15.2  & 2.7  & 16.5  & 12.1  & 79.5  & 49.3  & 58.1  & 23.0  & 28.4  & 8.1  & 49.0  & 35.8  & 28.5 & 0\\
      RangeNet++\cite{milioto2019rangenet++}   & 52.2  & \textbf{91.8}  & 75.2  & \underline{65.0}  & \textbf{27.8}  & 87.4  & 91.4 
      & 25.7  & 25.7  & 34.4  & 23.0  & 80.5  & 55.1  & 64.6  & 38.3  & 38.8  & 4.8  & 58.6  & 47.9  & 55.9 & 5\\
      3D-MiniNet\cite{alonso20203d}  & 55.8  & 91.6  & 74.5  & 64.2  & 25.4  & 89.4  & 90.5  & 28.5  & 42.3  & 42.1  & 29.4  & 82.8  & 60.8  & 66.7  & 47.8  & 44.1  & 14.5  & 60.8  & 48.0 
      & 56.6 & 0\\
      SqueezeSegV3\cite{xu2020squeezesegv3}  & 55.9  & \underline{91.7}  & 74.8  & 63.4  & \underline{26.4}  & 89.0  & 92.5  & 29.6  & 38.7  & 36.5  & 33.0  & 82.0  & 58.7  & 65.4  & 45.6  & 46.2  & 20.1  & 59.4  & 49.6 
      & \textbf{58.9} & 4\\
      PolarNet\cite{zhang2020polarnet}  & 54.3  & 90.8  & 74.4  & 61.7  & 21.7  & \underline{90.0}  & 93.8  & 22.9  & 40.2  & 30.1  & 28.5  & 84.0  & \underline{65.5}  & 67.8  & 43.2  & 40.2  & 5.6  & \underline{61.3}  &  51.8
      & \underline{57.5} & 4\\
      RandLA-Net\cite{hu2020randla}  & 55.9  & 90.5  & 74.0  & 61.8  & 24.5  & 89.7  & 94.2  & 43.9  & \textbf{47.4}  & 32.2  & 39.1  & 83.8  & 63.6  & 68.6  & 48.4  & 47.4  & 9.4  & 60.4  & 51.0 
      & 50.7 & 2\\
      CNN-LSTM\cite{wen2022hybrid}  & 56.9  & 90.7  & \underline{75.7}  & 23.3  & 17.6  & \underline{90.0}  & 92.6  & 48.6  & \underline{45.7}  & \textbf{49.6}  & 30.2  & \textbf{87.1}  & 60.8  & \textbf{75.4}  & \underline{53.8}  & \textbf{74.6}  & 9.2  & 51.3  & \textbf{63.9} 
      & 41.5  & \underline{14}\\
      BAAF-Net\cite{qiu2021semantic}  & \underline{59.9}  & 90.9  & 74.4  & 62.2  & 23.6  & 89.8  & \textbf{95.4}  & \underline{48.7}  & 31.8  & 35.5  & \textbf{46.7}  & 82.7  & 63.4  & 67.9  & 49.5  & 55.7  & \textbf{53.0}  & 60.8  & 53.7 
      & 52.0 & 8\\
      
      \midrule
      Proposed Method  & \textbf{61.1}  & 91.0  & \textbf{76.0}  & \textbf{65.5}  & 26.3  & \textbf{91.9}  & \underline{95.1}  & \textbf{48.8}  & 37.5  & \underline{43.3}  & \underline{46.4}  & \underline{84.2}  & \textbf{65.9}  & \underline{69.4}  & \textbf{54.6}  & \underline{58.8} & \underline{27.2}  & \textbf{66.3}  & \underline{55.5}  & 56.6 & \textbf{24}\\
  \bottomrule 
  \end{tabular}}
  \end{center}
  \end{table*}

\begin{table*}[h]
\caption{Ablation Results of the SIESEF-FusionNet Framework}
\label{table_3}
\begin{center}
\resizebox{14cm}{!}{
\begin{tabular}{lcccccccc}
\toprule
  Model  & ELSE  & SEAP  & Parameters(M)  & FLOPs(M) & Time(s)  & OA  & mIoU \\
  \midrule
    A1  & -  & -  & 4.7  & 28.0  & 61.6  & 96.9  & 80.8 \\
    A2  &\checkmark & -  & 5.4  & 32.2  & 62.0  & 97.1  & 81.8 \\
    A3  & -  &\checkmark  & 9.9  & 59.1  & 65.3 & 97.0  & 81.3 \\
    \midrule
    B1  & -  & -  & 5.0  & 26.4  & 61.1 & 96.3  & 79.4 \\
    B2  & \checkmark  & \checkmark  & 10.1  & 60.8  & 61.8 & 97.2 & 81.5 \\
    B3  & -  & -  & 5.0  & 29.7  & 112.5 &  97.1 & 80.9 \\
    B4  & \checkmark  & \checkmark  & 10.9  & 65.3  & 113.7 & 97.4  & 82.5 \\
    \midrule
  Proposed Method   & \checkmark  & \checkmark & 12.4 & 74.3 & 68.4  & 97.8  & 83.7\\
\bottomrule
\end{tabular}}
\end{center}
\end{table*}

\subsection{Evaluation on the SemanticKITTI Dataset}
Detailed quantitative comparisons between the proposed method and various baseline methods on the SemanticKITTI dataset are provided in Table \ref{table_2}. An $mIoU$ of 61.1\% is attained by the proposed method, outperforming other baseline models in 7 out of 19 classes, which are marked in bold in Table \ref{table_2}. It is noteworthy that superior performance is demonstrated by the proposed method not only in the recognition of small, densely scanned objects, such as trucks, pedestrians, and trunks, but also in the handling of large instances, such as buildings. In addition, suboptimal performance is achieved by the proposed method in eight classes, including cars, motorcycles, other-vehicles, vegetation, terrain, cyclists, motorcyclists, and poles, as indicated in italics in Table \ref{table_2}, with the $mIoU$ of cars and other-vehicles only 0.3\% lower than the best-performing method. It can be concluded that optimal or suboptimal segmentation performance is achieved by the proposed algorithm in nearly all semantic classes. Compared to existing point-based methods, including RandLA-Net and BAAF-Net, as well as projection-based methods such as SqueezeSegv3, RangeNet++, and PolarNet, a marked improvement in segmentation accuracy is demonstrated by the proposed method, with $mIoU$ increased by 1.2\% to 8.9\%. To more directly reflect the advantages of the proposed method, an $IoU$ or $mIoU$ achieving optimal performance is assigned a score of two points, while second-best performance is assigned a score of one point. The total scores of all methods are then calculated. As shown in Table \ref{table_2}, the highest score of 24 is achieved by the proposed method, which is 10 points higher than the second-ranked method and considerably higher than other baseline methods. The achievement of these results is attributed to the reverse residual feature aggregation module, through which spatial inter-correlations are embedded into semantic features, thereby alleviating the ambiguity of point cloud boundaries and improving the capability to achieve accurate semantic segmentation across varying object scales.

As illustrated in Fig. \ref{fig_4}, a comparison of the actual results between the proposed method and the results obtained from RandLA-Net and BAAF-Net on the validation set (Sequence 08) of the SemanticKITTI dataset is provided. The boundary segmentation of semantic classes, including persons and vegetation, trucks and vegetation, bicyclists and cars, as well as persons and roads, is clearly improved by the proposed method when compared to RandLA-Net and BAAF-Net. As shown in the second row of Fig. \ref{fig_4}(a), a person is missegmented as a bicyclist by RandLA-Net, while semantic boundary ambiguities between persons and vegetation are evident in BAAF-Net, resulting in a person being missegmented as vegetation. In the fourth row, boundary ambiguities between trucks and vegetation appear in both methods, leading to parts of the truck being missegmented as vegetation. In the second row of Fig. \ref{fig_4}(b), bicyclists are missegmented as cars to varying degrees by both methods. In the fourth row, a large portion of the person is missegmented as part of the road by RandLA-Net, while BAAF-Net entirely missegments the person as part of the road. These errors are caused by boundary ambiguities between different semantic classes, and the complete missegmentation of a pedestrian as part of the road in Fig. \ref{fig_4}(b) could seriously mislead decision-making in intelligent perception systems. In contrast, superior precision and smoother boundary segmentation between different semantic classes are achieved by the proposed method, whether for the person and vegetation or the truck and vegetation in Fig. \ref{fig_4}(a). Even for the bicyclist and person or any road in Fig. \ref{fig_4}(b), the results align closely with the ground truth. Within the complexity issues presented by the SemanticKITTI dataset, which encompasses traffic scenarios, ambiguities at the boundaries of different semantic classes are effectively alleviated by the proposed method, thereby enhancing the safety of autonomous driving systems and intelligent robot navigation.
\subsection{Ablation Studies}
To further analyze the effectiveness of the ELSE and SEAP modules, as well as their plug-and-play capability across different network architectures, extensive ablation experiments are conducted in this section. Due to limited opportunities for online validation on the SemanticKITTI dataset, ablation experiments are conducted on the Toronto3D dataset to thoroughly evaluate the performance of the proposed modules. To evaluate the scale and performance of various models, the total number of parameters, floating-point operations per second (FLOPs), and inference time are discussed. The inference time specifically refers to the duration required to process $10^{6}$ points.

The ablation results in Table \ref{table_3} demonstrate that both the ELSE and SEAP modules individually enhance $OA$ and $mIoU$. In comparison to the baseline model A1, an enhancement of 1.0\% in $mIoU$ has been recorded for model A2, emphasizing the critical role of ELSE in enhancing spatial information, which enriches spatial inter-correlation and facilitates an improvement in overall performance. Despite an observed increase of 4.2M in FLOPs, the inference time has been extended by only 0.4s. Although relative location information is solely relied upon by SEAP in the A3 model, an observed enhancement of 0.5\% in $mIoU$ is noted when compared to the A1 model, which does not incorporate SEAP. It is indicated that this enhancement results from the adaptive embedding of spatial information with semantic characteristics, leading to improved segmentation performance. The effectiveness of the proposed method is also validated by the ablation experiments conducted on variant models B1 to B3.

Both the ELSE and SEAP modules are integrated by the proposed method. Although the FLOPs of the proposed model are the highest among the seven variant models, reaching 74.3M, the inference time is assessed at a moderate level, extended by 6.8 seconds compared to the baseline model A1. Considering the comparison with all variant models, the $mIoU$ has been increased by 1.2\% to 4.3\%, while the $OA$ has been improved by 0.4\% to 1.5\%, indicating a marked improvement in segmentation performance, which renders the slight increase in inference time acceptable. The successful combination results from the adaptive embedding of the strong spatial inter-correlation, encoded by ELSE into semantic features through SEAP. The inter-correlation of semantic features is enhanced by this embedding, ambiguities in point cloud boundaries are mitigated, and segmentation precision is improved. The ablation experiments involving SEAP and ELSE across the distinct baseline models A1 and B1 have been conducted, thereby confirming the generalization capacity and flexibility of the proposed methodology. The visualization of segmentation results for the seven variant models and the proposed model is presented in Fig. \ref{fig_3}. Detailed analysis is provided in Section IV-C, where the exceptional performance of the proposed method is validated through the visual results.

\section{CONCLUSIONS}
A novel end-to-end method SIESEF-FusionNet for 3D LiDAR point cloud semantic segmentation is proposed in this study. The inter-correlation of spatial positional information is considerably improved by SIESEF-FusionNet through the use of inverse distance weighting and angle compensation. Problems with boundary ambiguity and vagueness in point clouds are effectively resolved by this approach, thereby enhancing the precision and reliability of data interpretation. A SEAP module is also developed to embed enhanced local spatial encodings into semantic features with enhancing their inter-correlation and addressing the inadequate integration of spatial positions with semantic features in existing pooling methods to achieve more accurate semantic segmentation. The performance of the proposed method is validated through comprehensive experiments on the Toronto3D and SemanticKITTI datasets. In both datasets, the best performance is achieved in both the building and fence classes, and the best or second-best performance is obtained in all the nature, vegetation, and car classes. The results of a series of ablation experiments indicate that considerable improvements are achieved by SIESEF-FusionNet in both segmentation performance and plug-and-play capability. Future work will focus on optimizing SIESEF-FusionNet to reduce memory consumption, enhance computational efficiency, and further improve segmentation accuracy.

\bibliographystyle{IEEEtran}
\bibliography{IEEEabrv,myrefs}

\end{document}